\documentclass[conference]{IEEEtran}
\IEEEoverridecommandlockouts
\usepackage{cite}
\usepackage{amsmath,amssymb,amsfonts}
\usepackage{algorithmic}
\usepackage{graphicx}
\usepackage{textcomp}
\usepackage{xcolor}
\usepackage{multirow}
\usepackage{tabularx}
\usepackage{booktabs} 
\usepackage{siunitx}
\usepackage[colorinlistoftodos]{todonotes}

\def\BibTeX{{\rm B\kern-.05em{\sc i\kern-.025em b}\kern-.08em
    T\kern-.1667em\lower.7ex\hbox{E}\kern-.125emX}}
\begin{document}

\title{Data Optimisation of Machine Learning Models for Smart Irrigation in Urban Parks\\}

\makeatletter
\newcommand{\linebreakand}{%
  \end{@IEEEauthorhalign}
  \hfill\mbox{}\par
  \mbox{}\hfill\begin{@IEEEauthorhalign}
}
\makeatother
\author{
    \IEEEauthorblockN{Nasser Ghadiri}
    \IEEEauthorblockA{
        \textit{School of Computer, Data and Mathematical Sciences} \\
        \textit{Western Sydney University, Australia}\\
        n.ghadiri@westernsydney.edu.au
    }
    \and
    \IEEEauthorblockN{Bahman Javadi}
    \IEEEauthorblockA{
        \textit{School of Computer, Data and Mathematical Sciences} \\
        \textit{Western Sydney University, Australia}\\
        b.javadi@westernsydney.edu.au
    }
    \linebreakand 
    \IEEEauthorblockN{Oliver Obst}
    \IEEEauthorblockA{
        \textit{School of Computer, Data and Mathematical Sciences} \\
        \textit{Western Sydney University, Australia}\\
        o.obst@westernsydney.edu.au
    }
    \and
    \IEEEauthorblockN{Sebastian Pfautsch}
    \IEEEauthorblockA{
        \textit{School of Social Sciences} \\
        \textit{Western Sydney University, Australia}\\
        s.pfautsch@westernsydney.edu.au
    }
}

\maketitle
\begin{abstract}
Urban environments face significant challenges due to climate change, including extreme heat, drought, and water scarcity, which impact public health, community well-being, and local economies. Effective management of these issues is crucial, particularly in areas like Sydney Olympic Park, which relies on one of Australia's largest irrigation systems. The Smart Irrigation Management for Parks and Cool Towns (SIMPaCT) project, initiated in 2021, leverages advanced technologies and machine learning models to optimize irrigation and induce physical cooling. This paper introduces two novel methods to enhance the efficiency of the SIMPaCT system's extensive sensor network and applied machine learning models.
The first method employs clustering of sensor time series data using K-shape and K-means algorithms to estimate readings from missing sensors, ensuring continuous and reliable data. This approach can detect anomalies, correct data sources, and identify and remove redundant sensors to reduce maintenance costs. The second method involves sequential data collection from different sensor locations using robotic systems, significantly reducing the need for high numbers of stationary sensors. Together, these methods aim to maintain accurate soil moisture predictions while optimizing sensor deployment and reducing maintenance costs, thereby enhancing the efficiency and effectiveness of the smart irrigation system.
Our evaluations demonstrate significant improvements in the efficiency and cost-effectiveness of soil moisture monitoring networks. The cluster-based replacement of missing sensors provides up to 5.4\% decrease in average error. The sequential sensor data collection as a robotic emulation shows 17.2\% and 2.1\% decrease in average error for circular and linear paths respectively. 
\end{abstract}


\section{Introduction}

The growing pressures of climate change, extreme heat, drought and water scarcity are all critical challenges for urban environments~\cite{8588668}. 
These challenges have enormous impacts on public health, community well-being, and local economies which are predicted to continue to intensify over the coming years~\cite{lee2023impacts}. With the growing interest in mitigating urban heat, the cooling effects of green infrastructure have been extensively studied and revealed the considerable air and land temperature reduction~\cite{jang2024street}. Public parks have been identified as effective urban infrastructure for cooling residential and commercial neighbourhoods and the Park Cool Island Effect is now well established in the literature \cite{declet2013creating,zhou2024exploring}.  

The Smart Irrigation Management for Parks and Cool Towns (SIMPaCT, see simpact-australia.com) project started in 2021 to develop smart technology to induce maximal cooling of the environment in Bicentennial Park. The area is a 42 ha public park located in Sydney’s Olympic Park precinct. SIMPaCT aimed at providing ideal soil moisture conditions to allow maximum transpiration cooling from plants during hot days while using the minimum irrigation water required to achieve a pronounced Park Cool Island Effect~\cite{pfautsch2023simpact}. The SIMPaCT system architecture including the Internet of Things (IoT) sensors, IoT platform and the Cloud platform is depicted in Figure~\ref{fig-arch}. The system has 202 soil moisture sensors, 50 ambient temperature and humidity sensors, and 13 weather stations which are connected via LoRaWAN to the IoT platform for data ingestion and data storage. Given the complexity in plant cover types, topography and depth of top soils, the sensor deployment strategy is designed to provide a high degree of spatial resolution in air temperature and soil moisture across the park. At the same time, the project team was determined to develop a lean and flexible digital architecture that can accommodate more or less IoT sensors in similar irrigation projects in the future. Hence, the scalability of the SIMPaCT solution is embedded at the core of the project. 

Today, the system ingests IoT and weather forecast data. These data are used by three hierarchically ordered analytical models that all generate irrigation commands for the park. The top model uses the data to simulate dynamic environmental changes across the park inside a digital twin. Here, geospatial models determine optimal irrigation schedules for 200 operational zones to maintain set ranges of soil moisture levels across the park. While producing highly accurate 3-day soil moisture forecasts and associated irrigation commands, this model is highly sensitive to missing data. If insufficient data is available, SIMPaCT falls back on a more robust model based on machine learning. This model adapts to short-term variations, relying only on the past two weeks of data. It does not become ‘smarter’ or more accurate over time. However, by focusing only on recent conditions and a seven-day forecast, the model constantly adjusts to seasonal variation. If all IoT sensors fail, for example, due to damage to the IoT gateways across the park, irrigation commands from a third model will be implemented. This model solely relies on air temperature and rainfall forecasting from an internet-based operator and predicts water losses in the park based on a crop evapotranspiration model (e.g., Penman-Monteith equation). While the redundancy built into SIMPaCT ensures continuity of the irrigation, the accuracy of the first and second models heavily depends on the reliability of the collected data from the sensor networks. Hence, finding a solution that further reduces this reliability is a sensible approach to harden the system against sensor failure.

\begin{figure}[t]
\centerline{\includegraphics[width=\columnwidth]{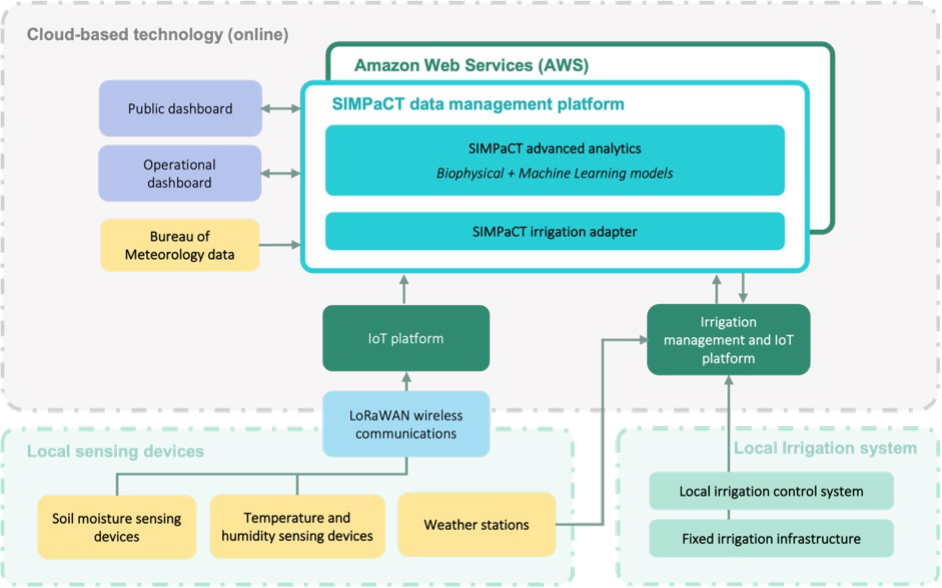}}
\caption{The SIMPaCT system architecture}
\label{fig-arch}
\end{figure}

Given the size and scale of SIMPaCT, sensor failure is an inevitable and critical challenge that will significantly impact the reliability, accuracy, and overall functionality of the irrigation system and ultimately plant health across the park. These failures can arise from various sources, including hardware malfunctions, environmental conditions, and battery issues~\cite{9174628}. Another challenge with the large sensor network is the cost and time required for the maintenance of sensors and the communication network. A key question is whether all of the sensors are required for accurate prediction. Some sensors could be redundant, as their time series may be similar to others \cite{prakaisak2022hydrological}, allowing for their removal. Recent research shows that reducing the number of IoT sensors or optimizing their number can maintain the same level of prediction accuracy with fewer sensors \cite{alam2020error, zhang2023autonomous, guyeux2019introducing}. Therefore, it is crucial to find ways to optimise (i.e., lower) the number of required sensors while maintaining the same level of prediction accuracy to ensure the sustainable operation of the SIMPaCT irrigation system.

This paper presents two novel methods for optimising a smart irrigation system equipped with soil moisture sensors. The first method involves clustering the sensor time series data using two distinct clustering algorithms: K-shape and Dynamic Time Warping (DTW)-based K-means. By grouping sensors with similar patterns, we can utilize the members of a cluster to estimate the time series of any missing sensor, ensuring continuity and reliability in the data. This approach not only addresses sensor failures but also optimises the number of sensors required for accurate monitoring by identifying and potentially removing redundant sensors.

The second method proposes a sequential data collection strategy using robots to visit different sensor locations for data collection. This method partitions the sensor network into groups, each assigned to a robot that sequentially collects data from the areas in its group at specified intervals. This approach significantly reduces the need for extensive sensor installations, lowering maintenance costs and enhancing the system's flexibility. The feasibility and cost-effectiveness of using robots for data collection are supported by recent studies~\cite{thayer2019bi, pulido2020kriging, loganathan2021agribot}, which highlighted their potential to reduce labour costs and improve data accuracy. 

These approaches aim to enhance the efficiency and effectiveness of the machine learning models for smart irrigation systems, promoting sustainable water management practices. The contributions of this work are as follows:
\begin{itemize}
    \item Development of clustering techniques (K-shape and DTW-based K-means) for real time soil moisture time series data to identify patterns and improve data reliability.
    \item Proposing a systematic method for estimating and replacing missing sensor data within clusters, reducing the need for additional sensor deployment.
    \item Implementing techniques for detecting anomalies in moisture sensor data, allowing for timely corrections and improving the accuracy and reliability of the sensor network.
    \item Proposing a novel data collection method using robotic systems to cover different locations, minimizing the dependency on a large number of fixed sensors and enhancing operational efficiency.
\end{itemize}

\section{Related Work}

The optimization of sensor networks in smart irrigation systems has been extensively studied. This section covers significant advancements in time series and IoT data clustering, handling missing data, applications of clustering in environmental domains, and the use of robots for sensor data collection. 

Time series clustering, particularly using algorithms like K-shape and DTW, has been widely studied across various applications. Paparrizos et al. ~\cite{paparrizos2015k} introduced the K-shape algorithm. This algorithm has been applied to various domains, which has been proven to be effective in grouping similar time series. Prakaisak et al.~\cite{prakaisak2022hydrological} applied clustering to hydrological time series from telemetry stations in Thailand, showing the practical benefits of these methods. Recent advancements in clustering methods for IoT data management have been significant. Alkawsi et al.~\cite{alkawsi2023towards} proposed a clustering algorithm for high-dimensional data streams to lower computational power in IoT systems. Similarly, Alam et al.~\cite{alam2020error} introduced an error-aware data clustering method for in-network data reduction in wireless sensor networks. 

Handling missing data is a critical aspect of managing sensor networks. Various imputation techniques are employed to estimate missing values based on the available data from other sensors. Lee et al.~\cite{lee2020clustering} used DTW for clustering water quality time series data, addressing missing data through robust clustering methods. 

The use of smart sensors in the environmental domain is a well-researched area. Soussi et al.~\cite{soussi2024sur} provided a comprehensive review of smart sensors and smart data in precision agriculture, highlighting the need for efficient data management and missing data handling in sensor networks. De et al.~\cite{de2023time} used time series clustering for analyzing carbon emissions from road transport, providing insights into emission patterns. Chen et al.~\cite{chen2024using} employed unsupervised learning to classify inlet water for industrial parks, ensuring stable water reuse designs.

The use of robots for sensor data collection in agricultural and environmental monitoring is gaining traction due to its feasibility and cost-effectiveness. Pulido et al.~\cite{pulido2020kriging} introduced kriging-based techniques for enhancing the accuracy of data collected by robots in agricultural settings. Loganathan et al.~\cite{loganathan2021agribot} developed Agribot, a robot designed specifically for agricultural applications, including sensor data collection and analysis, emphasizing its potential to reduce labour costs and improve data accuracy. These studies show the feasibility and cost-effectiveness of using robots for data collection, making them a valuable asset for optimizing smart irrigation systems.

The existing work is mainly focused on agriculture applications and many of them use synthetic or simulated data. In this paper, we used real data from a large-scale urban park which has different requirements in terms of irrigation. In addition, most of the clustering work addressed communication or computation overhead, while the objective in this paper is data reliability to improve the quality of the prediction model. Also, the existing robotic solutions are mainly focused on the design of robots and sensors or using the uniform deployment of sensors (e.g., in straight lines). In contrast, we are investigating using robots to collect the sensor data in a more complex environment.

\section{The Proposed Data Optimisation Techniques}

This section describes the proposed techniques for data optimisation of machine learning models in the smart irrigation system in Sydney Olympic Park. The approaches include clustering techniques for sensor time series data and a sequential data collection strategy using robotic sensing.

We use a cluster evaluation method to determine an approximate optimal number of clusters by evaluating the clustering quality within a specified range of \(k\) values. Within each cluster, we can identify sensors that are disconnected, have missing data, or should be removed due to high costs. To address these issues, we introduce a technique that emulates the missing or removed sensors by utilizing the values from other sensors within the same cluster.

\subsection{Using Sensor Time Series as Input for Clustering}
A time series is a sequence of data points collected at successive, typically uniform, intervals over time. Time series are generated by moisture sensors in our application. We apply time series clustering methods to group similar sensors based on their hourly moisture readings over a period of one or more months. These methods are flexible and can be applied to different time periods or types of sensors.

\subsection{Design Considerations and Methodology Selection}

Our primary goals were to enhance data reliability, reduce sensor network maintenance costs, and maintain accurate soil moisture predictions. We chose to explore both DTW-based K-means clustering and K-shape clustering:

\begin{itemize}
 \item DTW-based K-means: Selected for its ability to handle time series of varying lengths and its robustness to time warping.
 \item k-Shape Clustering: Chosen for its scale-invariance and ability to capture shape-based similarities in time series data.
\end{itemize}

We use the Silhouette score as our cluster evaluation metric due to its ability to provide insights into both cohesion within clusters and separation between clusters.

\subsection{DTW-based K-means Clustering Method}
\label{sec:dtw-def}

K-means clustering with DTW distance measure is used to partition time series into \(k\) clusters based on their similarity. DTW allows for non-linear warping of time series, accounting for stretches, compressions, and shifts commonly found in real-world data \cite{muller2007dynamic}.

\textbf{DTW distance measure:} The DTW distance between two time series \(X\) and \(Y\) is defined as:

\[
DTW(X, Y) = \min_P \sqrt{\sum_{(i, j) \in P} (x_i - y_j)^2}
\]
where \(P\) is the optimal warping path that aligns the elements of \(X\) and \(Y\).

The algorithm follows these steps:
\begin{enumerate}
    \item Initialize  \(k\) cluster centroids.
    \item Assign each time series to the nearest centroid based on DTW distance.
    \item Update cluster centroids.
    \item Repeat steps 2 and 3 until convergence.
\end{enumerate}

\subsection{k-Shape Clustering Method}
\label{sec:kshape-def}

K-shape clustering is designed for time series data and can handle missing data points~\cite{paparrizos2015k}. The shape-based distance between two time series \(X\) and \(Y\) is defined as:

\[
SBD(X, Y) = 1 - \max_w \frac{X \cdot Y_w}{\sqrt{X \cdot X} \sqrt{Y_w \cdot Y_w}}
\]
where \(Y_w\) is the optimally shifted version of \(Y\).
The algorithm follows similar steps to DTW-based K-means but uses shape-based distance and centroids.

\subsection{Silhouette Score for Evaluation of Clustering}
\label{sec:silh}

We use the Silhouette score \cite{rousseeuw1987silhouettes} to evaluate clustering quality and guide the selection of the near optimal number of clusters. For each time series \(X_i\), the silhouette score \(s(i)\) is computed as:

\[
s(i) = \frac{b(i) - a(i)}{\max{a(i), b(i)}}
\]
where \(a(i)\) is the average distance within the cluster and \(b(i)\) is the minimum average distance to other clusters. The average of silhouette scores overall clustered is calculated to provide an overall evaluation measure of clustering.

\subsection{Replacing the Time Series of a Missing Sensor}

To address missing sensor data, we replace it with the average from the most similar sensors within the same cluster. This approach ensures that the estimated value reflects the overall pattern and behaviour of the cluster.

\subsection{Sequential Data Collection Method for Robotic Sensing}

We propose partitioning the network of \(n_p\) sensors into \(n_s\) groups, each assigned to a robot for data collection. Each robot \(r_i\) visits sensors in its group \(g_j\) at intervals \(t_s\):

\[
r_i \to g_j, \quad \forall i \in \{1, 2, \ldots, n_s\}
\]

The total time for a complete round of data collection is:

\[
T_r = n_s \cdot t_s
\]

This approach reduces the number of fixed sensors required and optimizes the data collection process, leading to a more efficient and cost-effective irrigation system.

\section{Experimental Evaluation}
In this section we evaluate the proposed clustering methods, determining the number of clusters, using the results of clustering for sensor replacement, and emulation of robotic sequential data collection.

\subsection{Experimental Dataset}
We used collected data from the SIMPaCT sensor network in 2022 and 2023~\footnote{https://www.simpact-australia.com/}. This includes soil moisture data along with temperature and humidity data every 15 minutes, which are combined in the system to make predictions. Our experiments focus on moisture sensors, which are the primary type of sensors in the system. Other sensors like temperature and humidity sensors also contribute to prediction, so we maintained their values, and replacements were only targeted at selected moisture sensors.

\subsection{Using Silhouette Score for Determining k in Clustering}
As noted in Section \ref{sec:silh}, the silhouette score is an effective metric for evaluating the quality of clustering results. We computed the Silhouette score over a range of 4 to 40 clusters. In general, the score decreases with increasing k. 

However, it is not sufficient to rely solely on the average silhouette score when evaluating clustering results. While a higher average silhouette score may suggest better overall clustering quality, it does not account for the distribution of records among clusters. For instance, the K-shape clustering with $k=7$ yields the maximum average silhouette score in our experiments, but further inspection reveals that three out of the seven clusters are empty. The remaining clusters exhibit significant disparities in the number of records, with one cluster containing only 8 records while another contains 126 records. 

Therefore, it is essential to consider both the silhouette scores and the cluster occupancy when evaluating clustering performance. Based on this approach, and our experiments, we selected \(k=16\) for DTW-based K-means, and \(k=21\) for K-shape clustering in our experiments.

\subsection{Clustering of Moisture Sensor Time Series}
The input to the clustering algorithm can be any period of sensor time series data. To identify similar sensors and group them into clusters, the selected period should encompass as much variation as possible. 

While selecting one month of data for each sensor captures some variability, we chose to concatenate the values of two months, April and September 2023, to account for potential differences in sensor readings during warm and cold months. This approach ensures that the sensors grouped into clusters exhibit similar behaviour in both months. 

A sample of the results is shown in Figure~\ref{fig98}. Figure~\ref{fig98}a illustrates a cluster obtained using the K-shape clustering method. This cluster contains five sensors that exhibit nearly identical trends across both months. Figure~\ref{fig98}b displays a cluster from the DTW K-means clustering method, which includes 12 sensors. These sensors show similarities in the peaks and waveform patterns for both months.
\begin{figure*}[t]
\centerline{\includegraphics[width=\textwidth]{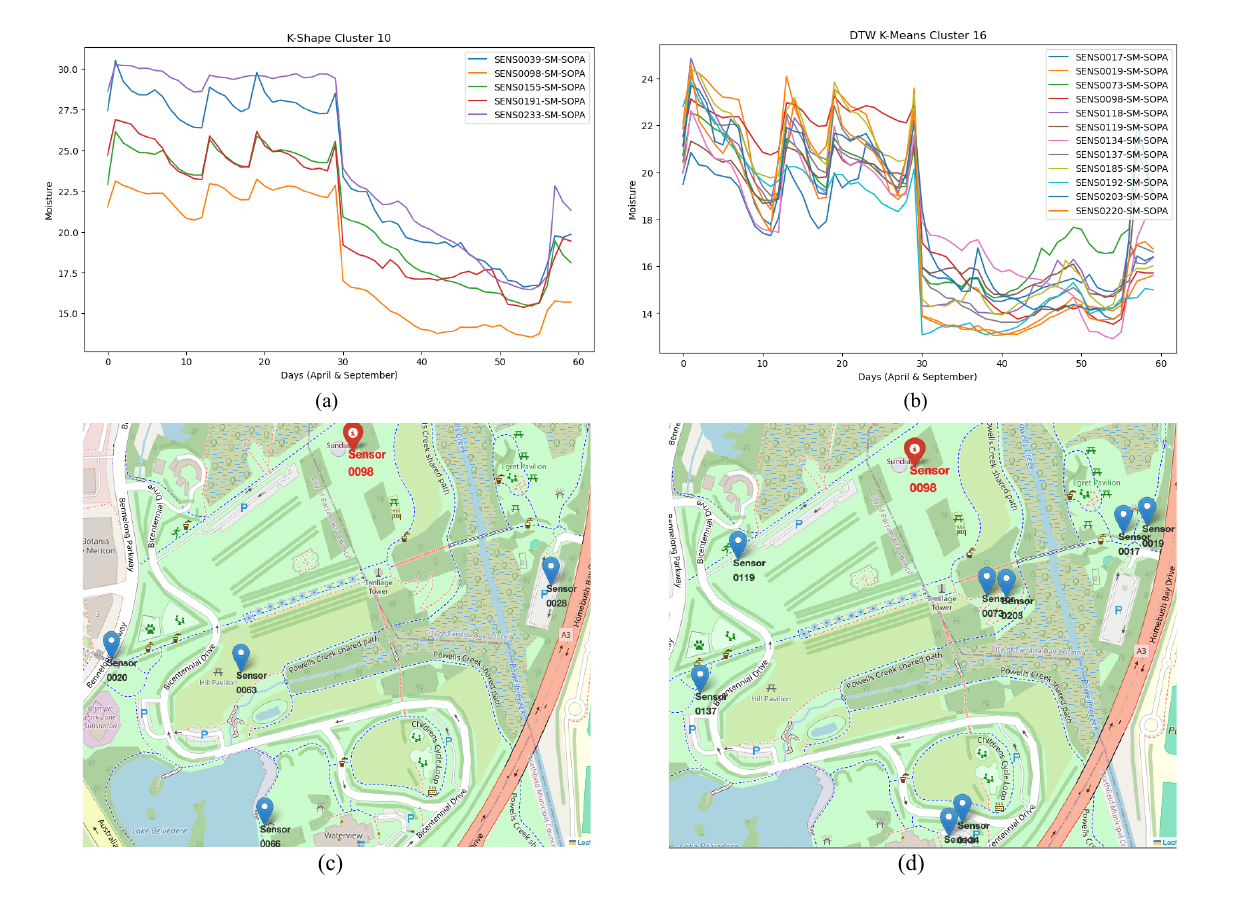}}
\caption{Clustering for sensor SENS0098-SM-SOPA replacement: (a) A cluster obtained using the K-shape clustering method, that contains five sensors showing nearly identical trends across both months. (b) A cluster from the DTW K-means clustering method, which includes 12 sensors, shows similarities in the peaks and waveform patterns for both months}
\label{fig98}
\end{figure*}

Figure~\ref{fig107} presents two more clusters from the same clustering methods. Figure~\ref{fig107}a illustrates another cluster obtained using the K-shape clustering method, which contains 10 sensors. These sensors exhibit nearly identical trends across both months, with the exception of one sensor that reports significantly different values in September. Figure~\ref{fig107}b displays another cluster from the DTW K-means clustering method, which includes six sensors. These sensors demonstrate similarities in the peaks and waveform patterns for both months.

\begin{figure*}[t]
\centerline{\includegraphics[width=\textwidth]{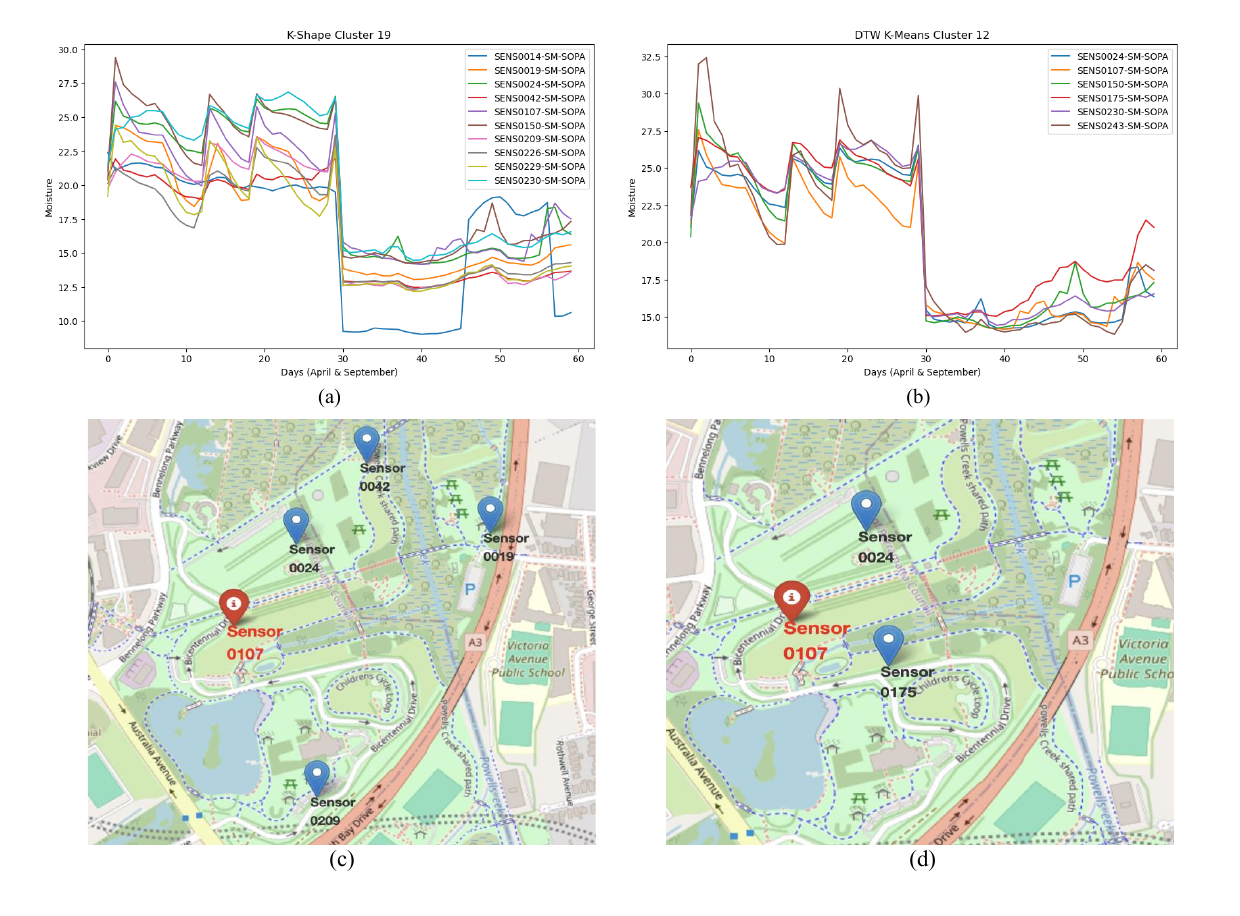}}
\caption{Clustering for sensor SENS0107-SM-SOPA replacement: (a) A cluster obtained using the K-shape, which contains 10 sensors that exhibit nearly identical trends across both months, with the exception of one sensor that reports significantly different values in September. (b) A cluster from the DTW K-means results, which includes six sensors. These sensors demonstrate similarities in the peaks and waveform patterns for both months.}
\label{fig107}
\end{figure*}

We utilized the selected clusters from both the K-shape and DTW-based K-means clustering methods to replace the values of sensors within each cluster. Specifically, we chose the moisture sensors SENS0098-SM-SOPA and SENS0107-SM-SOPA as target sensors. These sensors were selected due to their inclusion in the system's standard evaluation dataset, which provides a basis for our evaluations.

For each target sensor, we used other members of its cluster that existed in the evaluation dataset to calculate the average values for replacement. As shown in Figure~\ref{fig98}, sensors 19, 24, 42, and 209 were used to replace sensor 98 based on the K-shape clustering method. Using the DTW K-means clustering method, sensors 24 and 175 were selected to replace sensor 98.

\begin{figure*}[t]
\centerline{\includegraphics[width=\textwidth]{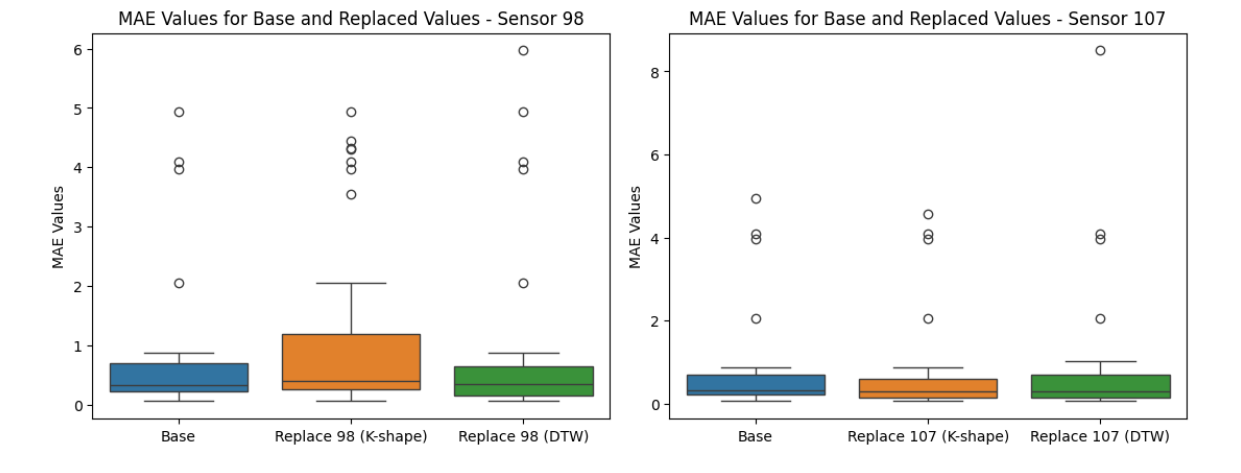}}
\caption{MAE box plot for using two clustering methods to replace SENS0098-SM-SOPA and SENS0107-SM-SOPA.}
\label{fig-mae-box}
\end{figure*}

\subsection{Results Of Sensor Replacement}

Mean Absolute Error (MAE) is a widely used metric for evaluating and comparing different prediction methods. In the SIMPaCT project, MAE has been utilized to select the best machine learning models for the production environment. Therefore, we use the same metric to assess the effectiveness of our proposed methods.

\subsubsection{Moisture Sensor 98}
The box plot for Sensor 98 in Figure~\ref{fig-mae-box} illustrates the MAE values for the base sensor data and the replaced values using the K-shape and DTW-based clustering methods.

\begin{itemize}
    \item \textbf{Base:} The base values represent the original MAE for Sensor 98 without any replacement. The median MAE is low, and the spread of errors is relatively small, indicating good performance of the original sensor data.
    \item \textbf{Replace 98 (K-shape):} The MAE values after replacing Sensor 98 using the K-shape clustering method show a higher median and a larger spread compared to the base values. This suggests that the replacement using the K-shape method introduces more variability and error in the predictions, potentially due to the diverse behaviour of the sensors in the cluster.
    \item \textbf{Replace 98 (DTW):} The MAE values after replacing Sensor 98 using the DTW-based clustering method have a median similar to the base values and a smaller spread compared to the K-shape method. This indicates that the DTW-based method provides a more accurate replacement with less variability, making it a more reliable approach for sensor replacement.
\end{itemize}

\subsubsection{Moisture Sensor 107}

The box plot for Sensor 107 in Figure~\ref{fig-mae-box} illustrates the MAE values for the base sensor data and the replaced values using the K-shape and DTW-based clustering methods.

\begin{itemize}
    \item \textbf{Base:} The base values represent the original MAE for Sensor 107 without any replacement. The median MAE is low, and the spread of errors is relatively small, indicating good performance of the original sensor data.
    \item \textbf{Replace 107 (K-shape):} The MAE values after replacing Sensor 107 using the K-shape clustering method show a slightly higher median compared to the base values, but with a similar spread. This suggests that the replacement using the K-shape method introduces some variability, but the overall error remains relatively low.
    \item \textbf{Replace 107 (DTW):} The MAE values after replacing Sensor 107 using the DTW-based clustering method have a median close to the base values and a smaller spread compared to the K-shape method. This indicates that the DTW-based method provides a more accurate replacement with less variability, making it a more reliable approach for sensor replacement.
\end{itemize}
\subsubsection{Overall Replacement Performance}

Table~\ref{tab1} presents the summary of MAE results for Sensors 98 and 107 under different clustering methods (K-shape and DTW). The percentage change indicates the improvement or degradation in MAE compared to the base values. For Sensor 98, the K-shape method significantly increased the MAE, indicating a less effective replacement, while the DTW method showed a smaller increase. For Sensor 107, the K-shape method slightly reduced the mean MAE, and the DTW method showed a moderate increase.

For both Sensor 98 and Sensor 107, the DTW-based clustering method appears to be more effective, resulting in MAE values closer to the base sensor data and lower variability. As mentioned in Section \ref{sec:dtw-def}, the DTW distance is robust to time warping, providing the ability to handle time series of varying lengths which is common in moisture sensors. The K-shape method provides scale-invariance as mentioned in Section \ref{sec:kshape-def}, but showed higher errors and greater variability, suggesting it is less suitable for accurate sensor replacement in this context. 

The result of clustering has also the potential to be used for sensor anomaly detection. When visualizing some clusters, for example, the K-shape cluster number 19 in  Figure~\ref{fig107}(a), the sensors appear similar for the first month. However, one sensor (sensor 14 shown in blue) deviates significantly from the rest of the cluster in the second month. This deviation could be normal or abnormal. Investigating such anomalies helps prevent potential failures and reduces forecast errors.

\subsection{Robot Emulation Results}

We selected two groups of size \( n_s = 6 \), one resembling a linear path and the other a circular path. We assumed the moisture value at each sensor location is measured and collected by a robot every \( t_s = 120 \) minutes. Therefore, we used only the sensor values recorded at intervals of two hours.

The MAE values for the robot emulation are shown in Figure~\ref{figrob}. For the linear path, the  MAE is almost the same as the base values. As shown in Table \ref{tab1}, MAE is decreased by 2.1\%, suggesting that the linear path configuration can provide accurate moisture measurements. For the circular path, Figure~\ref{figrob} shows that the median MAE for the circular path is slightly lower than the base values, with a slightly higher spread. The mean of MAE reported in Table \ref{tab1} in this case is decreased by 17.2\%. This indicates that while the circular path configuration also provides accurate measurements, there is slightly higher variability compared to the linear path. Overall, both the linear and circular path configurations for the robot emulation demonstrate effectiveness in maintaining accurate moisture measurements with low error rates.

The integration of robots into data collection processes can be hindered by technical limitations, such as battery lifetime, maintenance, and operational challenges in different weather conditions. As technology evolves, overcoming these challenges will likely lead to increasingly sophisticated and effective robotic data collection systems, ultimately transforming applications and research fields.

\begin{figure*}[t]
\centerline{\includegraphics[width=\textwidth]{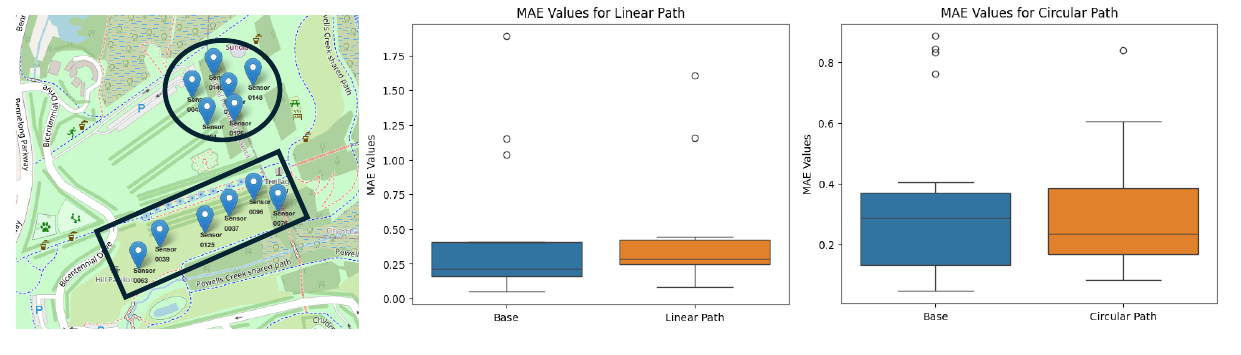}}
\caption{Selected areas for robot emulation and the resulting MAE values for each area: median MAE for the circular path is slightly lower than the base values, with a slightly higher spread}
\label{figrob}
\end{figure*}

\begin{table}[htbp]
\caption{MAE Values for Sensors 98 and 107, and Robotic Paths}
\centering
\begin{tabular}{|l|l|p{1.35cm}|p{1.35cm}|r|}
\hline
\textbf{Method} & \textbf{Condition} & \textbf{Base MAE} & \textbf{MAE} & \textbf{Change} \\ 
\hline
\multirow[t]{3}{*}{Clustering} & S98/K-shape  & \multirow{2}{*}{$0.77 \pm 1.21$} & $1.27 \pm 1.62$ & $\uparrow$ \SI{64.9}\percent \\ 
\cline{2-2} \cline{4-5} 
                            & S98/DTW      &                       & $ 0.93 \pm 1.51  $& $\uparrow$ \SI{20.8} \percent  \\ 
\cline{2-5} 
                            & S107/K-shape & \multirow{2}{*}{$0.77 \pm 1.21$} & $0.73 \pm 1.18 $& $\downarrow$ \,  \SI{5.4}\percent \\ 
\cline{2-2} \cline{4-5} 
                            & S107/DTW     &                       & $0.85 \pm 1.68 $& $\uparrow$ \SI{10.3} \percent  \\ 
\hline
\multirow[t]{2}{*}{Robotic}    & Linear & $0.47 \pm 0.55$  & $0.46 \pm 0.43$ & $\downarrow$ \, \SI{2.1}\percent \\ 
\cline{2-5} 
                            & Circular  & $0.34 \pm 0.28$ &$ 0.29 \pm 0.20$ & $\downarrow$ \SI{17.2}\percent \\ 
\hline
\end{tabular}
\label{tab1}
\end{table}

\section{Conclusion and future work}
Urban parks play an increasingly critical role as key environments for cooling, biodiversity and public health in densifying and warming cities. Many regions of the world experience increasing seasonal scarcity of rainfall. Hence, optimizing the water requirements of public parks while keeping them green and vibrant will enhance the resilience of local populations against heat impacts and encourage the use of smart city technology.

In this paper, we presented how smart irrigation systems for urban green infrastructure can be optimised to lower operational costs and increase their efficiency. The results of our experiments demonstrate that both methods, the clustering of time series data and robot-operated data collection, effectively maintain accurate soil moisture predictions while optimizing the number of sensors and reducing maintenance costs. The DTW-based clustering method, in particular, showed superior performance in accurately replacing sensor values with minimal error and variability.

In the future, the implementation of these optimization techniques can bring significant benefits to existing smart irrigation systems, particularly a reduction of operational costs for park managers. We plan to evaluate the proposed technique for other urban parks in Australia under different weather conditions. Given the flexible approach to the core digital architecture of the project and specifically its capacity to respond to changes in the spatial resolution of IoT data, we are confident that the SIMPaCT solution can be applied to irrigation projects with greater or less complexity.


\section*{Acknowledgment}

We would like to thank the SIMPaCT team who helped to access the dataset for this paper.

\bibliography{SIMPaCT-refs, IEEEabrv,mybibfile}

\end{document}